%% file: main.tex
\pdfoutput=1
\documentclass[11pt]{article}
\usepackage[final]{acl}
\usepackage{times}
\usepackage{latexsym}
\usepackage[T1]{fontenc}
\usepackage[utf8]{inputenc}
\usepackage{microtype}
\usepackage{inconsolata}
\usepackage{graphicx}
\usepackage{xspace}
\input{math_commands.tex}

\usepackage{hyperref}
\usepackage{booktabs}
\usepackage{multirow,makecell}
\usepackage{url}
\usepackage{subcaption}
\usepackage{xcolor}
\usepackage[most]{tcolorbox}
\providecommand{\imagUp}[1]{\textcolor{green!55!black}{$\uparrow$\,#1}}

\usepackage[table]{xcolor}
\definecolor{lightpurple}{HTML}{F0ECF5}

\definecolor{findinggray}{RGB}{242,242,242}
\newtcolorbox{findingbox}{
  colback=findinggray,
  colframe=black,
  boxrule=0.8pt,
  arc=8pt,
  left=6pt,
  right=6pt,
  top=2pt,
  bottom=2pt,
  enhanced,
  drop shadow={black!18!white},
}

\title{Representing the Unseen: Rethinking Visual Imagination \\ for Spatial Reasoning}

\author{Zengqi Zhao\thanks{Equal contribution.} \quad
Zun Wang\footnotemark[1] \quad
Shoubin Yu \quad
Mohit Bansal \quad
Yue Zhang \\
UNC Chapel Hill \\[0.5em]
\href{https://imagirep.github.io/}{\texttt{https://imagirep.github.io/}}
}

\newcommand{\bench}{\textsc{ImagiRep}\xspace}

\newcommand{\method}{\textsc{GEARS}\xspace}

\begin{document}

\maketitle

\input{sections/abstract}
\input{sections/intro}

\input{sections/related_work}

\input{sections/benchmark}

\input{sections/controlled_study}

\input{sections/method}

\input{sections/conclusion}

\bibliography{references}

\clearpage
\appendix

\input{sections/appendix}

\end{document}

%% file: math_commands.tex
\usepackage{amsmath,amsfonts,bm}

\def\eqref#1{equation~\ref{#1}}

\def\1{\bm{1}}

\DeclareMathAlphabet{\mathsfit}{\encodingdefault}{\sfdefault}{m}{sl}
\SetMathAlphabet{\mathsfit}{bold}{\encodingdefault}{\sfdefault}{bx}{n}

%% file: sections/abstract.tex
\begin{abstract}
Visual world models increasingly support spatial reasoning by generating RGB views of unobserved states. However, RGB has become the default interface largely because it is well suited for visual generation, rather than necessarily being the most effective representation for downstream reasoning. Different spatial questions may benefit from different representations of the same imagined state: top-down geometry can facilitate global distance comparisons, while panoramic views provide broader context for reasoning about open space.
To isolate the role of representation in spatial imagination, we introduce \textbf{\bench{}}, a simulator-grounded benchmark of 922 questions spanning seven question types across two settings, Viewpoint Imagination and Cross-view Integration, providing oracle RGB and six alternative representations (e.g., depth, panoramic views, and top-down geometry) of the same unobserved state.
This controlled comparison reveals that even accurate imagined RGB does not guarantee accurate spatial reasoning, and different questions benefit from different representations. 
Similar trends hold for representations derived from world-model-estimated states. Notably, estimated RGB alone can mislead spatial reasoning, while alternative representations retain complementary benefits.
Building on these findings, we introduce \method{}, a lightweight test-time strategy for adaptive representation selection, outperforming existing spatial imagination methods and showing that reasoning about unseen states requires more than generating RGB views.

\end{abstract}

%% file: sections/intro.tex
\section{Introduction}

Recent spatial reasoning systems increasingly use visual world models to imagine what lies beyond the current observation. Many spatial questions involve changes in viewpoint, such as what becomes visible after a turn or how object relations change after moving. Given a movement or hypothetical action, these systems generate future RGB views and feed them back to a vision-language model for reasoning~\cite{yang2025mindjourney,yu2026avic,zhu2026thinking,yuan2026fast}. These generated views provide explicit evidence about an unobserved state, complementing what the model can infer from its initial observations.

\begin{figure*}
    \centering
    \includegraphics[width=\linewidth]{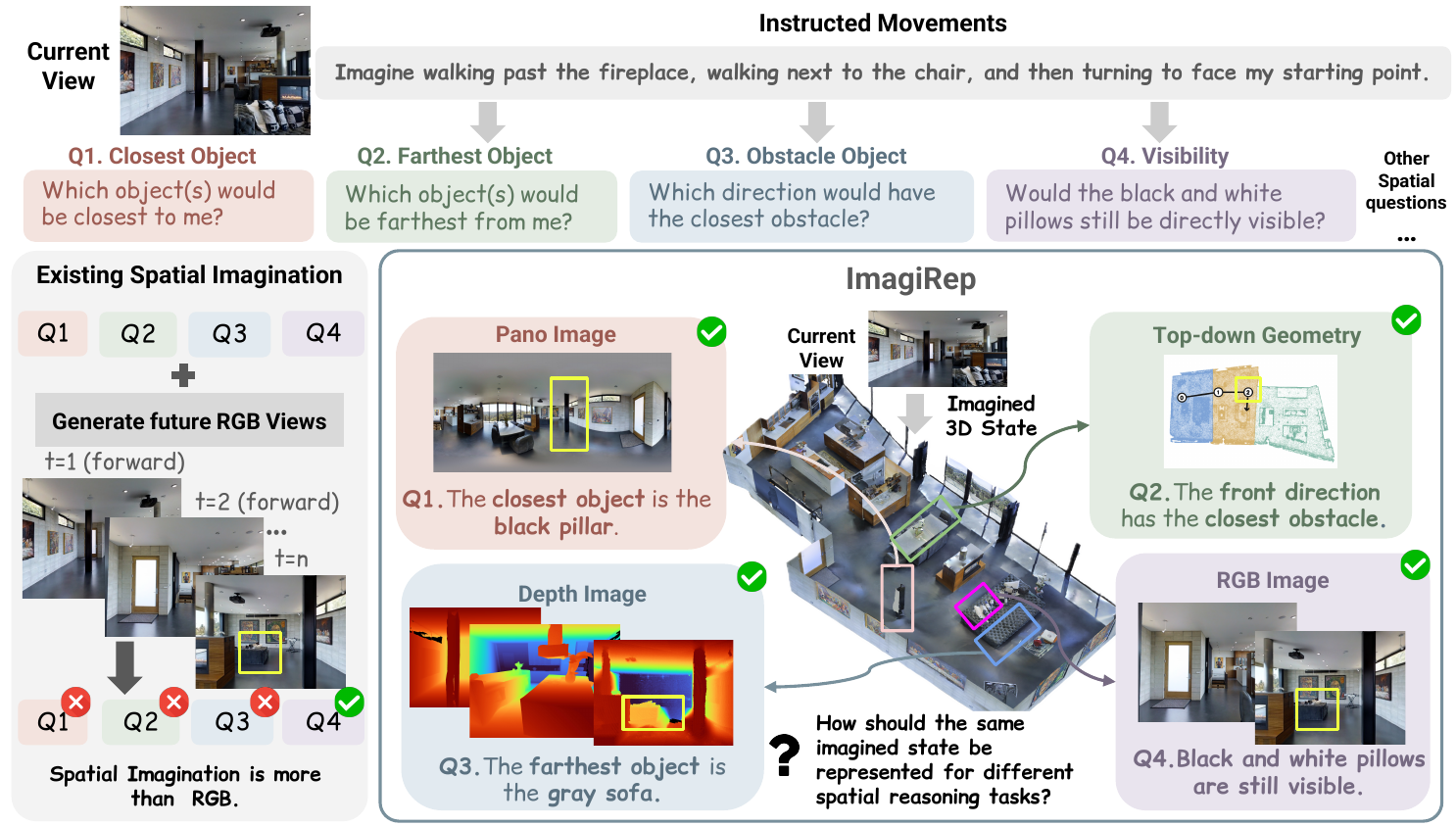}
    \caption{
    Spatial imagination is more than generating future RGB views. The same imagined state may support different spatial questions (e.g., top-down
geometry helps global distance comparisons; panoramic views facilitate reasoning about open space), yet existing approaches typically expose it to the reasoner only through RGB. \bench{} provides controlled access to the same underlying state through multiple representations, allowing us to study how the choice of representation affects downstream spatial reasoning.}
    \label{fig:teaser}
\end{figure*}

RGB is a natural representation for generation because it preserves rich visual appearance and is the standard output of modern visual world models~\cite{agarwal2025cosmos,du2023learning,bar2025navigation,zhou2025stable}. Consequently, existing spatial imagination methods largely treat RGB as the default interface between world models and downstream reasoners~\cite{yu2026avic, yang2025mindjourney}. However, this choice is driven by how imagined states are generated, rather than by which representation best supports spatial reasoning across different question types. 
As illustrated in Fig.~\ref{fig:teaser}, the same imagined spatial state can support diverse questions about visibility, distance, and free space, where representations such as depth and panorama can make distance cues and surrounding spatial context more explicit than RGB alone.
This raises a fundamental question: \textit{should all spatial reasoning tasks rely on imagined RGB views, or could alternative representations better support different reasoning needs?}

To understand how different representations affect spatial reasoning, we need to compare them under the same underlying spatial state.
Therefore, we first introduce \textbf{\bench{}}, a simulator-grounded benchmark for systematically studying how different representations of imagined spatial states affect reasoning. \bench{} includes two complementary settings: \textit{Viewpoint Imagination} (Fig.~\ref{fig:benchmark}(a)), which requires reasoning about spatial relations after an instructed movement, and \textit{Cross-view Integration} (Fig.~\ref{fig:benchmark}(b)), which requires combining information across partial observations. 
Using ground-truth scene geometry and camera poses, we obtain oracle RGB and six alternative representations (\textbf{Depth}, \textbf{Surface Normals}, \textbf{Panorama}, \textbf{Top-down Geometry}, \textbf{Spatial Text}, and \textbf{RGB Video}). 
To examine how representation utility varies across reasoning needs, we design seven question categories across two settings, covering directional reasoning, nearest/farthest-object comparisons, visibility, and spatial ordering. For example, open-space reasoning may benefit from panoramic views, while global distance comparisons may benefit from top-down geometry. By keeping the underlying spatial state and question fixed, \bench{} enables controlled comparisons of how different representations support spatial reasoning.

Our findings reveal a gap between the accuracy of imagined views and their usefulness for spatial reasoning, a pattern observed across both GPT-6.1-sol and Gemini 3.7 Flash. For example, with GPT-6.1-sol, adding oracle endpoint RGB reduces accuracy on Farthest Object questions from 68.4\% to 58.2\%. Yet top-down geometry raises the corresponding score to 75.5\%, showing that how spatial information is presented matters.
More broadly, different representations provide complementary benefits across models and questions, with no single representation consistently performing best. 
Beyond oracle representations, we further investigate a realistic setting where world models estimate unobserved spatial states and provide corresponding representations for reasoning. Interestingly, representations derived from these estimated states can still improve spatial reasoning, even when estimated RGB views alone degrade performance. 
Together, these findings suggest that effective spatial imagination depends not only on the accuracy of imagined views, but also on how spatial information is represented and understood by the reasoner.

Motivated by these findings, we next introduce \method{} (\textbf{G}ated \textbf{E}vidence and \textbf{A}daptive \textbf{R}epresentation \textbf{S}election) (shown in Fig.~\ref{fig:gears}), a lightweight test-time strategy that adaptively determines when to use imagined evidence and which representation to select.
Rather than assuming that imagined RGB is always beneficial, \method{} first examines the agreement between predictions based on observed evidence and those incorporating estimated RGB.
When the predictions disagree, it adaptively selects among observed evidence and estimated representations to support spatial reasoning.
\method{} achieves 61.4\% accuracy on Viewpoint Imagination and 77.7\% on Cross-view Integration, outperforming other spatial imagination methods.
Nevertheless, a substantial gap remains compared with oracle-assisted results, highlighting the need for world models to not only estimate unobserved states more accurately, but also provide spatial information that reasoners can effectively use.

%% file: sections/related_work.tex
\section{Related Work}

\noindent\textbf{Visual Spatial Reasoning.}
Spatial reasoning in vision-language models has been studied across a broad range of abilities, from relative position, depth, and orientation in single images~\citep{chen2024spatialvlm,fu2024blink, ma20253dsrbench, zhang2024spartun3d, zhang2026seeing} to perspective taking and reasoning under ego and object motion~\citep{ray2024sat,li2026viewspatial}, cross-view correspondence~\citep{allangles,xu2025mmsibench}, mental modeling from limited views~\citep{mindcube}, and spatial memory and situated awareness in video~\citep{yang2025thinking,yang2026cambrian,li2026sawbench}.
While these benchmarks evaluate diverse spatial reasoning abilities, they do not systematically examine how different representations of the same unobserved state affect reasoning. In contrast, our \textbf{\bench{}} provides multiple oracle representations of the same simulator-grounded target state, enabling controlled comparisons while keeping the spatial state and question fixed.

\noindent\textbf{Spatial Imagination with World Models.}
Action- and camera-conditioned video models can now render plausible unseen views~\citep{agarwal2025cosmos,bar2025navigation,zhou2025stable,parker2025genie,wang2026epic,wang2026anchorweave}, and several works feed these views back to an agent or reasoner~\citep{lu2024generative,huang2025vista,yang2025mindjourney,yuan2026fast,hwang2026spatio}.
AVIC~\citep{yu2026avic} and Astra~\citep{zhu2026thinking} further explore selective world-model use through controlled views or camera motions.
However, these methods primarily rely on imagined RGB views, despite emerging world models that can also generate depth, surface normals, and camera geometry~\citep{zhu2025aether}.
Yet the ability to generate these representations does not establish whether they are more useful for downstream spatial reasoning.
Our work addresses this by systematically comparing representations of the same unobserved state under oracle and estimated imagination, and exploring adaptive representation selection at test time.

%% file: sections/benchmark.tex
\section{\textbf{\bench{}}: A Controlled Benchmark for Spatial Imagination}
\label{sec:benchmark}

\begin{figure*}
    \centering
    \includegraphics[width=\linewidth]{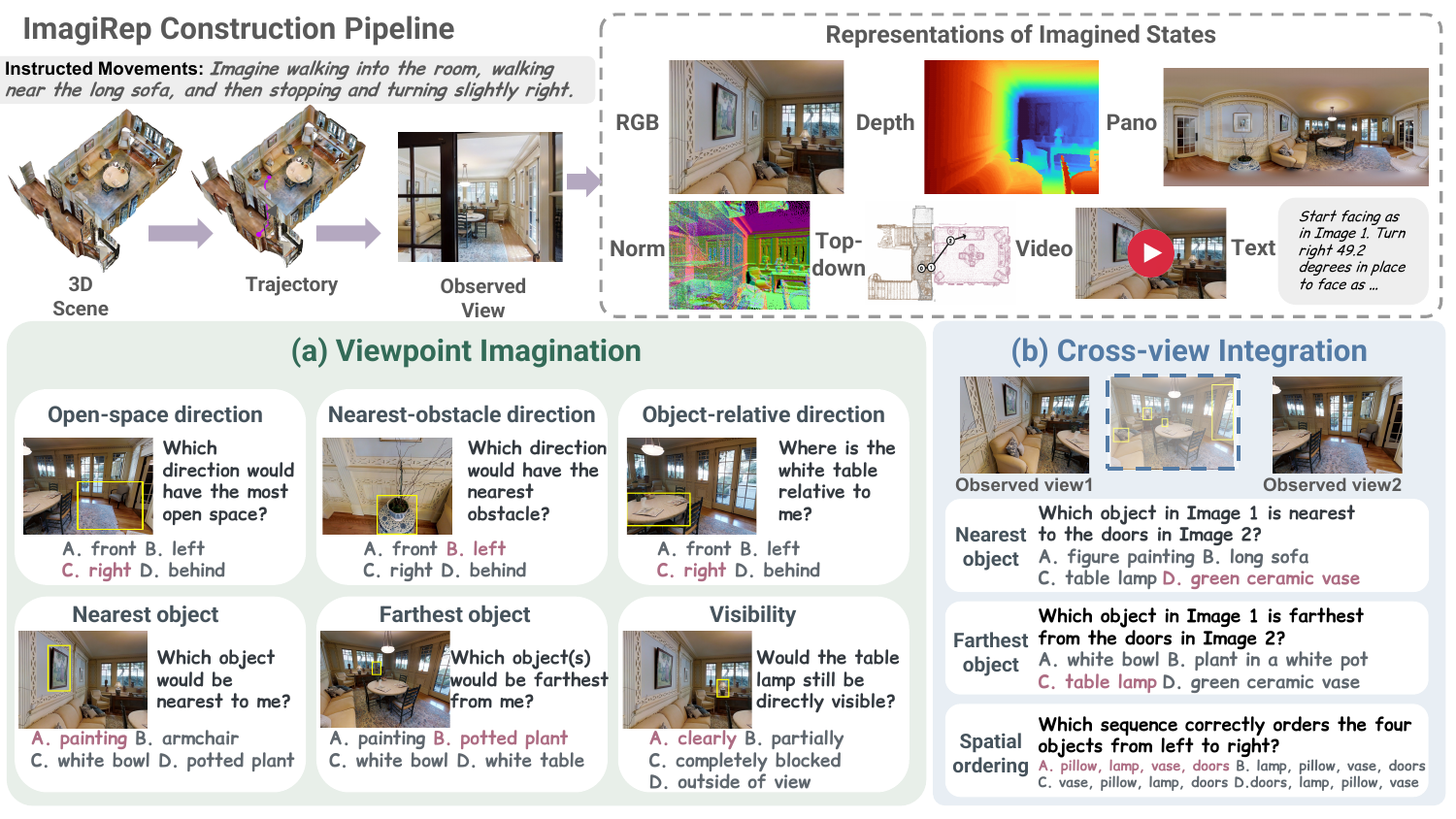}
    \caption{Overview of \textbf{\bench{}}. The top left illustrates benchmark construction, including a 3D scene, movement instructions, and observed view. The top right shows oracle RGB and six alternative representations of the same imagined state. The lower panels present example questions and ground-truth answers for the Viewpoint Imagination and Cross-view Integration question types.}
    \label{fig:benchmark}
\end{figure*}

\subsection{Benchmark Overview and Design Principles}

Spatial cognition involves reasoning about environments beyond the immediately observed view, including updating spatial relations after movement, integrating information across viewpoints, and reasoning within different spatial reference frames~\cite{rieser1989access}. 
Prior research on spatial cognition has studied these processes and the role of spatial representations in supporting them~\cite{tversky2003structures,burgess2006spatial, mindcube,ray2024sat}. 
Inspired by these distinctions, we design \textbf{\bench{}} to evaluate spatial reasoning over unobserved states through diverse questions and controlled representation inputs.

Recent benchmarks~\cite{ray2024sat, xu2025mmsibench,mindcube} evaluate these spatial reasoning capabilities through questions involving perspective changes, object relations, and imagined movements. However, these benchmarks are not designed to provide controlled access to the underlying spatial states across different representations. This makes it difficult to isolate the effect of representation choice from errors in estimating or generating the imagined state. 
To address this limitation, we construct \textbf{\bench{}} using Matterport3D~\cite{chang2017matterport3d}, which provides 3D scene geometry and camera poses. 
This allows us to obtain different representations of the same ground-truth spatial state while keeping the underlying scene and question fixed.
\textbf{\bench{}} follows three design principles: \textbf{(1) Reasoning over unobserved spatial states}, with questions involving viewpoint changes and spatial integration across partial observations; \textbf{(2) Diverse spatial reasoning tasks}, covering directions, distances, visibility, and object ordering to probe different spatial information needs; and \textbf{(3) Controlled representation comparisons}, where the same underlying spatial state and question are held fixed while the representation of imagined evidence is varied.

To achieve this, as shown in Fig.~\ref{fig:benchmark}, we sample movement routes and groups of views from Matterport3D scenes for Viewpoint Imagination and Cross-view Integration, respectively. Human annotators write spatial questions based on the resulting observations and movement instructions. Using the corresponding scene geometry and camera poses, we obtain oracle RGB and six alternative representations of imagined evidence. The question, observed inputs, and answer options remain fixed across representation conditions, allowing us to isolate the effect of representation choice.

\subsection{Spatial Imagination Tasks}

Reasoning about unobserved spatial states presents two complementary challenges: updating spatial relations after a change in viewpoint and integrating spatial information distributed across partial observations. We design two settings to study representation utility under these different conditions.

\noindent\textbf{(1) Viewpoint imagination} requires reasoning about a new viewpoint after an instructed movement, where spatial relations must be updated relative to the observer's new position and orientation. Given an initial RGB observation, the model receives a movement instruction specifying a route and final heading, and answers questions about the resulting viewpoint. We sample straight and turning routes from the navigable connectivity graph, spanning one to three edges and approximately 0.5--9 meters. We favor routes connecting different rooms and endpoints surrounded by recognizable objects to support diverse spatial questions.
We design six question types that probe different spatial relations: \textit{Open-space Direction}, \textit{Nearest-obstacle Direction}, and \textit{Object-relative Direction} for directional reasoning; \textit{Nearest Object} and \textit{Farthest Object} for distance comparison; and \textit{Visibility} for determining whether the starting area remains visible after movement.

\begin{table*}[!t]
\centering

\resizebox{\textwidth}{!}{%
\begin{tabular}{llrrrrrrrrc}
\toprule
& & & & \multicolumn{6}{c}{Endpoint RGB + Additional Representation} & Oracle \\
\cmidrule(lr){5-10}
Task & Question Type & Observed View & + Endpoint RGB & Top-down & Text & Video & Depth & Normals & Panorama & Selection \\
\midrule
\multirow{7}{*}{\makecell[l]{Viewpoint\\Imagination}}
 & Open-space Direction & 60.4 & 57.1 & \textbf{74.7} & 69.2 & 65.9 & 64.8 & 61.5 & \underline{73.6} & 89.0 \\
 & Nearest-obstacle Direction & 38.2 & 49.1 & 61.8 & 58.2 & 60.0 & \underline{63.6} & 58.2 & \textbf{65.5} & 81.8 \\
 & Visibility & 39.4 & 64.4 & 73.1 & 71.2 & \underline{76.0} & 73.1 & 75.0 & \textbf{76.9} & 80.8 \\
 & Nearest Object & 51.0 & 64.4 & \underline{76.9} & 73.1 & 76.0 & \underline{76.9} & 71.2 & \textbf{80.8} & 87.5 \\
 & Farthest Object & 68.4 & 58.2 & \textbf{75.5} & 67.3 & \underline{72.4} & 70.4 & 67.3 & 71.4 & 87.8 \\
 & Object-relative Direction & 70.0 & 82.9 & 88.2 & 87.6 & \underline{89.4} & 86.5 & 85.9 & \textbf{91.8} & 92.4 \\
 \cmidrule(l){2-11}
 & All & 57.2 & 66.1 & \underline{77.5} & 74.0 & 76.2 & 74.9 & 72.7 & \textbf{79.3} & 87.5 \\
\midrule
\multirow{4}{*}{\makecell[l]{Cross-view\\Integration}}
 & Nearest Object & 56.8 & 74.1 & 79.0 & 79.0 & 77.8 & \textbf{84.0} & \underline{80.2} & 76.5 & 87.7 \\
 & Farthest Object & 59.8 & 54.9 & \textbf{70.7} & 67.1 & \underline{69.5} & \underline{69.5} & 59.8 & 64.6 & 85.4 \\
 & Spatial Ordering & 79.6 & 89.1 & 92.0 & \textbf{95.6} & \underline{94.2} & \underline{94.2} & 92.7 & \textbf{95.6} & 97.8 \\
 \cmidrule(l){2-11}
 & All & 68.0 & 75.7 & 82.7 & \underline{83.3} & 83.0 & \textbf{84.7} & 80.3 & 82.0 & 91.7 \\
\bottomrule
\end{tabular}
}

\caption{Complementary benefits  of \textbf{oracle representations}. Oracle selection counts a question as correct if the endpoint RGB alone or any one of the additional representations answers it correctly. Videos are provided as 16 uniformly sampled frames. Best and second-best results are marked in bold and underlined, respectively.}
\label{tab:task_breakdown_gpt}
\end{table*}

\noindent\textbf{(2) Cross-view integration} examines how spatial information distributed across partial observations can be combined to reason about an unobserved viewpoint. When relevant objects appear in different views, their spatial relations are not directly observable in any single image, requiring the model to integrate information across viewpoints. 
To study this, we construct groups of two or three partially overlapping RGB observations captured from different views.
We consider \textit{Cross-view Nearest Object} and \textit{Cross-view Farthest Object}, which follow the same reasoning tasks as in Viewpoint Imagination but require comparing objects across different views. We additionally introduce \textit{Spatial Ordering}, which asks the model to determine the ordering of objects relative to a specified viewpoint.

\begin{table}[!tb]
\centering

\resizebox{\linewidth}{!}{%
\begin{tabular}{lrrrr}
\toprule
& \multicolumn{2}{c}{Gemini 3.7 Flash} & \multicolumn{2}{c}{GPT-6.1-sol} \\
\cmidrule(lr){2-3} \cmidrule(lr){4-5}
Input & \makecell{Viewpoint\\Imagination} & \makecell{Cross-view\\Integration} & \makecell{Viewpoint\\Imagination} & \makecell{Cross-view\\Integration} \\
\midrule
Random Selection          & 25.0 & 25.0 & 25.0 & 25.0 \\
Question Only             & 31.7 & 29.7 & 31.0   & 31.0   \\
Observed View(s)          & 53.7 & 66.3 & 57.2 & 68.0 \\
\midrule
+ Oracle Endpoint RGB     & \textbf{63.7} & \textbf{78.7} & \textbf{66.1} & \textbf{75.7} \\
\bottomrule
\end{tabular}
}
\caption{Accuracy (\%) of Gemini 3.7 Flash and GPT-6.1-sol under different input conditions.}
\label{tab:oracle_imagination}
\end{table}

\subsection{Representations of Imagined States}
For each question, we identify the target viewpoint in Matterport3D and render the corresponding oracle RGB image using its ground-truth camera pose. Leveraging the simulator's 3D geometry and camera information, we further obtain six additional representations: \textit{Depth}, \textit{Surface Normals}, \textit{Panorama}, \textit{Top-down Geometry}, \textit{Spatial Text}, and \textit{Video} (Fig.~\ref{fig:benchmark}). Depth and surface normals are rendered from the target viewpoint, while panoramas and top-down maps provide broader spatial context. Spatial text describes camera movements derived from ground-truth poses, and videos capture movement along the route or transitions between observed views. All representations are derived from the same underlying 3D scene, enabling controlled comparisons of their utility for spatial reasoning.

\subsection{Dataset Statistics and Quality Control}
\textbf{\bench{}} contains 922 human-annotated spatial questions, including 622 Viewpoint Imagination questions from 200 routes across 73 scenes and 300 Cross-view Integration questions from 143 view groups across 21 scenes (Table~\ref{tab:question_counts} in Appendix~\ref{app:benchmark}). Human annotators write the questions and identify the correct answers from four candidate options based on the corresponding scenes and viewpoints. All examples undergo manual review to ensure that the spatial relations are inferable from the provided observations and movement instructions. We additionally check rendering quality, camera-pose alignment, and annotation consistency, remove duplicate routes, and randomize answer options to reduce potential biases manually.

%% file: sections/controlled_study.tex
\section{Evaluating Imagined Evidence for Spatial Reasoning}
\label{sec:oracle_analysis}

\begin{table*}[!t]
\centering

\resizebox{\textwidth}{!}{%
\begin{tabular}{llrrrrrrrrc}
\toprule
& & & & \multicolumn{6}{c}{+ Estimated Representation} & Per-Type \\
\cmidrule(lr){5-10}
Task & Question Type & Observed View & + Est.\ Endpoint RGB & Top-down & Text & Video & Depth & Normals & Panorama &  Best \\
\midrule
\multirow{7}{*}{\makecell[l]{Viewpoint\\Imagination}}
 & Open-space Direction       & 60.4 & 52.7 & \underline{63.7} & 60.4 & 62.6 & 57.1 & 57.1 & \textbf{70.3} & 70.3 \\
 & Nearest-obstacle Direction & 38.2 & 45.5 & 56.4 & 54.5 & \textbf{61.8} & 54.5 & 52.7 & \underline{60.0} & 61.8 \\
 & Visibility                 & 39.4 & 47.1 & 50.0 & 51.9 & \underline{54.8} & 52.9 & 52.9 & \textbf{56.7} & 56.7 \\
 & Nearest Object             & 51.0 & 49.0 & 57.7 & 56.7 & \underline{59.6} & \textbf{61.5} & 58.7 & \underline{59.6} & 61.5 \\
 & Farthest Object            & 68.4 & 53.1 & 62.2 & \underline{69.4} & \textbf{70.4} & 66.3 & 67.3 & 63.3 & 70.4 \\
 & Object-relative Direction  & 70.0 & 67.6 & 72.4 & \underline{73.5} & 72.4 & 71.2 & 71.2 & \textbf{75.3} & 75.3 \\
 \cmidrule(l){2-11}
 & All                        & 57.2 & 54.7 & 61.9 & 62.9 & \underline{64.6} & 62.2 & 61.7 & \textbf{65.6} & 67.2 \\
\bottomrule
\end{tabular}
}
\caption{Complementary benefits of \textbf{estimated representations} on Viewpoint Imagination. Reported scores retain correct predictions from estimated endpoint RGB and include additional questions answered correctly with each representation. Per-Type Best reports the highest score across representations for each question type. Best and second-best results among the six representations are marked in bold and underlined, respectively. }
\label{tab:task_breakdown_estimated}
\end{table*}

\begin{table}[!tb]
\centering

\resizebox{\linewidth}{!}{%
\begin{tabular}{lrrr}
\toprule
Question Type & Observed View & + Est.\ RGB & + Est.\ Depth \\
\midrule
Nearest Object  & 56.8 & \underline{74.1} & \textbf{79.0} \\
Farthest Object & \underline{59.8} & 54.9 & \textbf{65.9} \\
Spatial Ordering           & 79.6 & \underline{89.8} & \textbf{93.4} \\
\midrule
All                        & 68.0 & \underline{76.0} & \textbf{82.0} \\
\bottomrule
\end{tabular}
}
\caption{Complementary benefits of estimated representations on Cross-view Integration. Best and second-best results are marked in bold and underlined, respectively.}
\label{tab:task_breakdown_estimated_cv}
\end{table}

In this section, we evaluate how imagined evidence affects spatial reasoning. We first test whether oracle endpoint RGB improves performance (Sec.~\ref{sec:oracle_future}), then examine whether alternative oracle representations provide further gains (Sec.~\ref{sec:representation_selection}), and finally assess whether these benefits persist with estimated imagination (Sec.~\ref{sec:estimated_imagination}).

\subsection{Experimental Setup}
\label{sec:oracle_setup}

\noindent\textbf{Models and evaluation metric.}
We evaluate two multimodal reasoning models, GPT-6.1-sol~\citep{openai2026gpt61sol} and Gemini 3.7 Flash~\citep{googledeepmind2026gemini37flash}, using high reasoning effort. Since GPT-6.1-sol achieves higher oracle-selection accuracy across both settings, we use it as the primary reasoner for estimated imagination and test-time experiments. 
Both models receive identical questions, answer options, and representation-specific inputs in the oracle comparisons, and are instructed to select one of four answer choices (A--D). We report accuracy (\%) as the percentage of correctly answered questions.

\noindent\textbf{Input conditions.}
We first establish three conditions: question-only, observed view(s), and observed view(s) augmented with oracle endpoint RGB. We then examine six alternative representations of the same target state: top-down geometry, spatial text, video, depth, surface normals, and panorama. Each representation is provided alongside the observed view(s) and endpoint RGB, while the question, answer options, and underlying spatial state remain fixed. We additionally evaluate estimated representations to study the effect of imperfect imagination.

\noindent\textbf{Estimated representations.}
We use HY-World 2.0~\citep{hyworld2026} for Viewpoint Imagination, generating a 3D scene from the initial observation and following a trajectory planned from the movement instruction. Endpoint RGB and the alternative visual representations are rendered from this generated scene, while spatial text describes the planned camera movements. For Cross-view Integration, we use Stable Virtual Camera (SEVA)~\citep{zhou2025stable} to synthesize endpoint RGB from the observed views and consider depth estimated from the synthesized RGB as the additional representation. Implementation details are provided in Appendix~\ref{app:estimation}.

\subsection{Does Oracle Imagination Improve Spatial Reasoning?}
\label{sec:oracle_future}

We first examine whether providing oracle imagined views improves spatial reasoning over unobserved states. As shown in Table~\ref{tab:oracle_imagination}, Question-only performance is close to the random baseline, suggesting that language priors alone provide limited information for answering these spatial questions. 
Providing observed views substantially improves 20+\% absolute accuracy for both models and settings, suggesting that VLMs can already infer some spatial relations without explicit imagined evidence. Interestingly, both models achieve lower accuracy on Viewpoint Imagination than on Cross-view Integration, suggesting that reasoning about spatial changes after movement is more challenging than integrating information across observed views.
Adding an oracle endpoint RGB further improves accuracy by around 10\% across both models and settings. Nevertheless, substantial errors remain even with ground-truth endpoint views, suggesting that accurate imagined RGB alone is insufficient for reliable spatial reasoning.

The benefits of oracle RGB vary considerably across question types. 
As shown in Tables ~\ref{tab:task_breakdown_gpt} and ~\ref{tab:task_breakdown_gemini}, both models achieve substantial improvements on visibility and object-relative direction questions, while performance on farthest-object reasoning remains unchanged or even decreases. For example, GPT-6.1-sol drops from 68.4\% to 58.2\% on Farthest Object despite receiving the ground-truth endpoint view. 
This observation is consistent with AVIC~\cite{yu2026avic}, which highlights the importance of selectively invoking imagination rather than assuming that additional imagined views are always beneficial.

\begin{findingbox}
\textbf{Finding 1:}
Oracle imagined RGB improves spatial reasoning overall, but can degrade performance on certain question types, such as Farthest Object, showing that accurate visual evidence does not guarantee better reasoning.
\end{findingbox}

\begin{figure*}
    \centering
    \includegraphics[width=\linewidth]{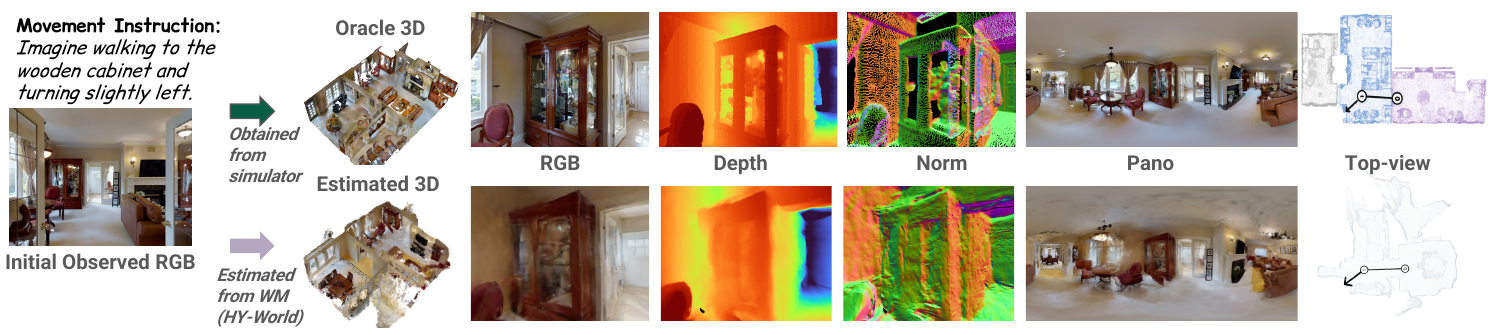}
\caption{Example comparison of oracle and estimated spatial representations for \textbf{Viewpoint Imagination}. Oracle representations are from ground-truth 3D scenes; estimated representations are obtained from HY-World 2.0.}
    \label{fig:vis_example}
\end{figure*}

\subsection{Can Alternative Representations Improve Reasoning Beyond RGB?}
\label{sec:representation_selection}

We next investigate whether alternative representations can provide useful spatial information beyond oracle endpoint RGB. As shown in Table~\ref{tab:task_breakdown_gpt}, all six representations recover some errors made with oracle RGB across both reasoning settings. For GPT-6.1-sol, panorama achieves the highest overall recovery score on Viewpoint Imagination (79.3\% vs. 66.1\% with RGB), while depth performs best on Cross-view Integration (84.7\% vs. 75.7\%). Similar improvements are observed with Gemini 3.7 Flash, as shown in Table~\ref{tab:task_breakdown_gemini} in Appendix~\ref{app:gemini} where top-down geometry performs best on Viewpoint Imagination (73.0\% vs.\ 63.7\% with RGB) and video on Cross-view Integration (85.3\% vs.\ 78.7\%). 
These results suggest that even when the imagined RGB is accurate, alternative representations can provide complementary evidence that the reasoner does not fully extract from RGB alone.

\begin{figure}
    \centering
\includegraphics[width=\linewidth]{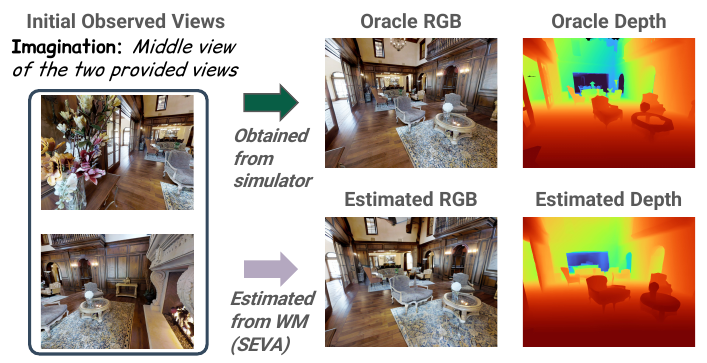}
   \caption{Example comparison of oracle and estimated representations for \textbf{Cross-view Integration}. Estimated RGB is synthesized by SEVA, and estimated depth is predicted from the synthesized RGB.}
\label{fig:cross_estimation}
\end{figure}

Importantly, the benefits of different representations depend on the spatial information required by the question. Top-down geometry performs best on Farthest Object questions (75.5\% vs. 58.2\% with RGB), suggesting that explicit scene layout helps reason about objects distributed across a larger area. Panorama improves Open-space Direction from 57.1\% to 73.6\%, where a wider field of view provides more information about the surrounding space. In contrast, depth performs best on Cross-view Nearest Object questions (84.0\% vs. 74.1\% with RGB), where explicit distance cues may facilitate comparisons across viewpoints. These patterns suggest different representations make different spatial relationships easier to reason about by making relevant spatial cues more explicit.

This complementarity also explains why no single representation consistently performs best. Oracle selection across representations achieves 87.5\% on Viewpoint Imagination and 91.7\% on Cross-view Integration, exceeding the best fixed representations (79.3\% and 84.7\%, respectively). Although this represents an upper bound based on ground-truth correctness, it highlights the potential of selecting representations according to the spatial evidence needed for each question.

\begin{findingbox}
\textbf{Finding 2:} Alternative representations provide complementary spatial evidence beyond RGB, with their benefits varying across question types. No single representation consistently performs best.
\end{findingbox}

\subsection{Do Representation Benefits Persist under Imperfect Imagination?}
\label{sec:estimated_imagination}

So far, our analysis has focused on oracle representations derived from ground-truth 3D scenes. We now investigate whether their complementary benefits persist when the imagined state must be estimated from observed views. 
Unlike the oracle setting, estimated representations may contain errors in scene geometry and viewpoint placement (see examples in Fig.~\ref{fig:vis_example} for Viewpoint Imagination and Fig.~\ref{fig:cross_estimation} for Cross-view Integration). We first evaluate estimated endpoint RGB, then examine whether additional estimated representations can recover its remaining errors. More implementation details are provided in Appendix~\ref{app:estimation}.

As shown in Table~\ref{tab:task_breakdown_estimated}, estimated endpoint RGB reduces accuracy from 57.2\% to 54.7\% on Viewpoint Imagination, potentially due to errors in scene reconstruction and instruction-conditioned viewpoint estimation. Nevertheless, all six alternative representations provide complementary evidence, yielding potential error-recovery scores of 61.7-65.6\%. 
Their effectiveness varies across question types: panorama performs best on Open-space Direction (70.3\%), while depth performs best on Nearest Object (61.5\%). Selecting the best representation per question type further increases the score to 67.2\%. In contrast, estimated RGB improves Cross-view Integration from 68.0\% to 76.0\%. This difference may reflect the greater difficulty of reconstructing a 3D scene and estimating an instructed movement endpoint from a single observation in Viewpoint Imagination.  Estimated depth further increases the score to 82.0\% (Table~\ref{tab:task_breakdown_estimated_cv}). Together, these results suggest that representation complementarity persists even under imperfect imagination.

\begin{findingbox}
\textbf{Finding 3:} Even when estimated RGB alone does not help, alternative representations derived from the same estimated scene still improve spatial reasoning.
\end{findingbox}

\begin{figure}
    \centering
    \includegraphics[width=\linewidth]{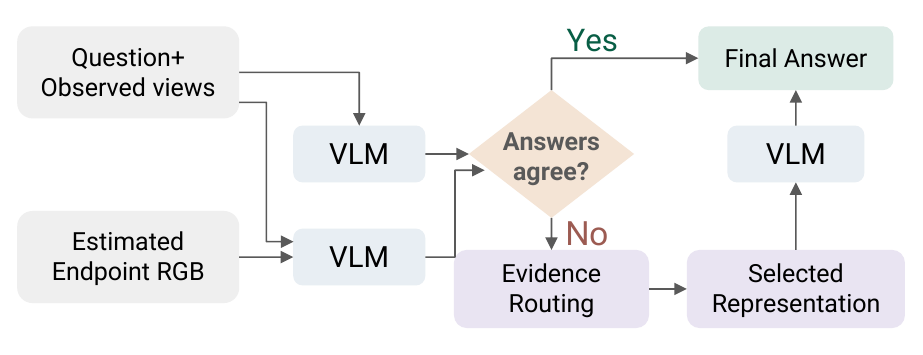}
  \caption{Overview of \method{}, combining prediction-agreement gating with adaptive representation selection.}
    \label{fig:gears}
\end{figure}

%% file: sections/method.tex
\section{Our Method: Gated Evidence and Adaptive Representation Selection (\method)}
\label{sec:method_benchmark}

Our findings in Sec.~\ref{sec:oracle_analysis} reveal two challenges for practical spatial imagination: estimated RGB does not always improve reasoning, and different representations provide complementary spatial information. These observations motivate adaptive decisions about when to use imagined evidence and which representation to provide.

We therefore introduce \method{}, as shown in Fig.~\ref{fig:gears}, a lightweight test-time strategy that decides when to use imagined evidence and which representation to provide. A gating stage keeps the answer when predictions with and without the estimated endpoint RGB agree, while a routing stage selects the evidence that best suits the spatial information required by the question. For Viewpoint Imagination, we consider all six alternative representations, while for Cross-view Integration, we consider estimated depth as the additional representation.

We compare GEARS against two existing spatial imagination methods, MindJourney~\cite{yang2025mindjourney} and AVIC~\cite{yu2026avic}. For a consistent comparison, all methods use GPT-6.1-sol as the final reasoner and are evaluated on the same benchmark questions. 
Implementation details, an ablation of the two stages, and baseline configurations are provided in Appendix~\ref{app:baseline}.
As shown in Table~\ref{tab:wm_baselines}, GEARS also achieves the highest accuracy on most question types (61.4\% accuracy on Viewpoint Imagination and 77.7\% on Cross-view Integration). 
These results demonstrate that the complementary benefits identified in our controlled analysis can be translated into practical improvements through adaptive representation selection.

\begin{table}[!t]
\centering

\small
\setlength{\tabcolsep}{3pt}
\begin{tabular*}{\linewidth}{@{\extracolsep{\fill}}lccc@{}}
\toprule
Question Type & MindJourney & AVIC & \textbf{GEARS} \\
\midrule

\rowcolor{lightpurple}\multicolumn{4}{@{}l}{\textbf{Viewpoint Imagination}} \\
Open-space Direction       & 53.8 & 51.6 & \textbf{63.7} \\
Nearest-obstacle Direction & 52.7 & 49.1 & \textbf{54.5} \\
Visibility                 & 45.2 & 46.2 & \textbf{48.1} \\
Nearest Object             & 53.8 & 51.0 & \textbf{55.8} \\
Farthest Object            & \textbf{69.4} & 63.3 & 68.4 \\
Object-relative Direction  & 67.1 & 62.4 & \textbf{70.0} \\
\cmidrule(lr){1-4}
\textbf{Overall}           & 58.4 & 55.1 & \textbf{61.4} \\
\midrule
\addlinespace[5pt]
\rowcolor{lightpurple}
\multicolumn{4}{@{}l}{\textbf{Cross-view Integration}} \\
Nearest Object             & 61.7 & 59.3 & \textbf{74.1} \\
Farthest Object            & \textbf{62.2} & 51.2 & 61.0 \\
Spatial Ordering           & 81.8 & 85.4 & \textbf{89.8} \\
\cmidrule(lr){1-4}
\textbf{Overall}           & 71.0 & 69.0 & \textbf{77.7} \\
\bottomrule
\end{tabular*}
\caption{Comparison with test-time spatial imagination methods on \bench{}. Accuracy (\%) is reported for each question type, with the best results in bold.}
\label{tab:wm_baselines}
\end{table}

%% file: sections/conclusion.tex
\section{Conclusion}

We introduce \bench{}, a controlled benchmark for studying how representations of imagined spatial states affect reasoning. Our findings reveal that accurate imagined RGB does not guarantee effective reasoning, while alternative representations offer complementary benefits even under imperfect imagination. Building on these insights, \method{} improves test-time spatial reasoning. More broadly, our study calls for a shift in world model development from merely generating accurate views to providing spatial representations that enable effective reasoning about the unseen world.

\section{Limitations}

Our study has several limitations. First, \bench{} focuses on static indoor scenes and selected spatial reasoning tasks, leaving dynamic environments and more complex interactions unexplored. Second, our estimated representation analysis relies on a limited set of world models, and not all representations are evaluated across both settings. Finally, despite the improvements achieved by \method{}, a substantial gap remains compared with oracle-assisted results, suggesting that reliable estimation and selection of imagined spatial evidence remain open challenges. Future work could extend our framework to more diverse environments, world models, and reasoning tasks.

\section{Acknowledgements}
This work was supported by NSF-AI Engage Institute DRL-2112635, NSF-CAREER Award 1846185, ARO Award W911NF2110220, ONR Grant N00014-23-1-2356, and a Google Faculty Award. The views contained in this article are those of the authors and not of the funding agency.

%% file: sections/appendix.tex
\section{Appendix}
\makeatletter
\setlength{\@fptop}{0pt}\setlength{\@fpsep}{12pt}\setlength{\@fpbot}{0pt plus 1fil}
\setlength{\@dblfptop}{0pt}\setlength{\@dblfpsep}{12pt}\setlength{\@dblfpbot}{0pt plus 1fil}
\makeatother
\renewcommand{\dbltopfraction}{0.85}
\setcounter{dbltopnumber}{2}
\subsection{Details of \bench{}}
\label{app:benchmark}

\paragraph{Additional statistics.}
Of the 200 Viewpoint Imagination trajectories, 46 involve one move, 139 involve two moves and 15 involve three moves between navigable viewpoints. The median total length of a trajectory is 4.2\,m. 52 trajectories include a turn of at least $45^\circ$, and the rest are approximately straight. Each trajectory carries one to six questions, and 257 questions (41\%) ask the observer to turn back toward the starting point at the end of the movement. All observed views in a Cross-view Integration view group are rendered from the same camera position with a $15^\circ$ downward pitch, and adjacent views are $98.5^\circ$ apart.

\begin{figure*}[!t]
    \centering
    \includegraphics[width=\linewidth]{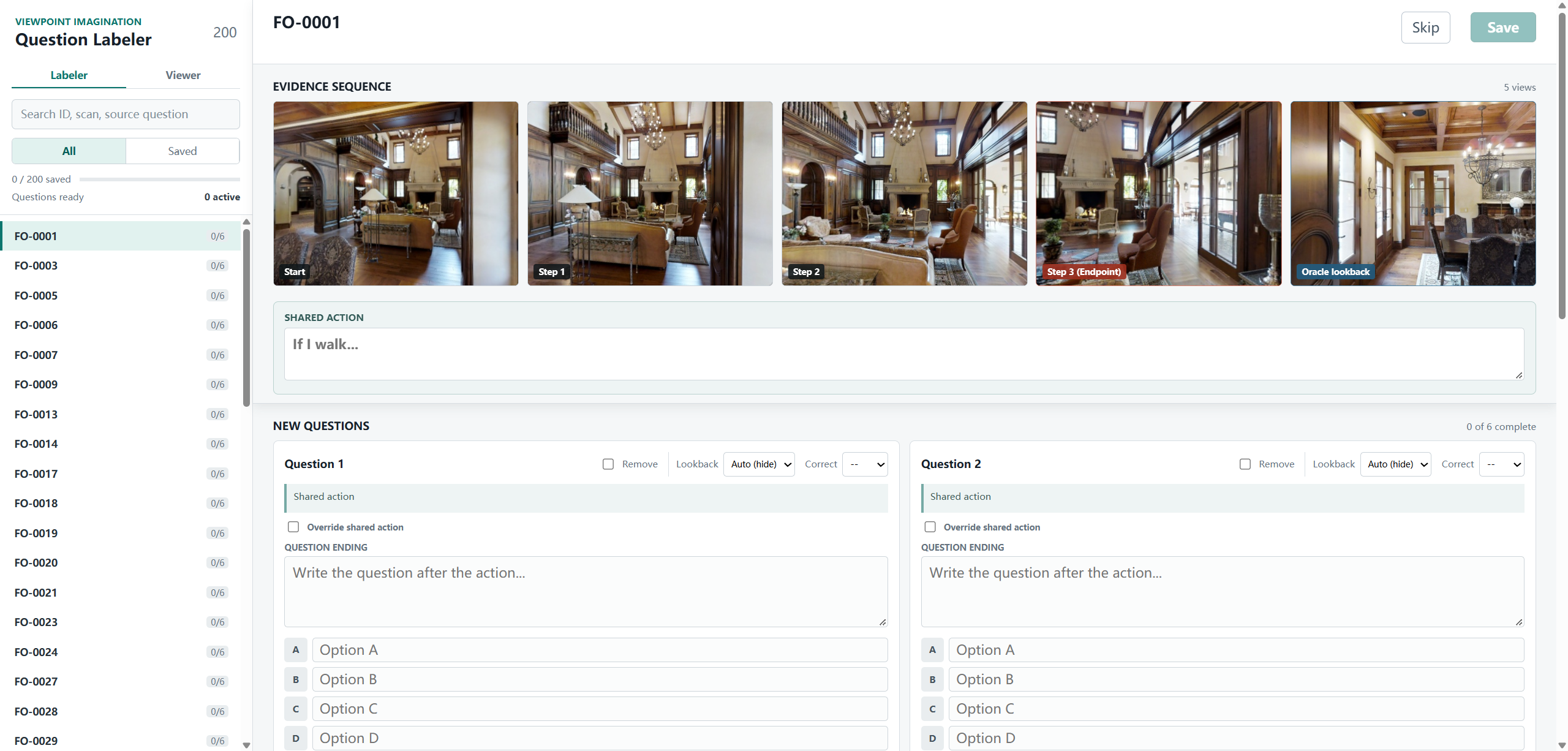}
    \caption{Annotation interface for Viewpoint Imagination.}
    \label{fig:labeler}
\end{figure*}

\paragraph{Annotation process.}
Figure~\ref{fig:labeler} shows the web tool used to annotate Viewpoint Imagination.
For each trajectory, annotators first review the views rendered by the simulator along the trajectory and write a shared movement description that begins every question on that trajectory. They then complete a question for each question type, write the question ending and four answer options, and mark the correct answer. When a question asks the observer to turn back, a look-back view rendered at the endpoint and facing the start serves as its endpoint evidence.

\subsection{Results on Oracle Representations with Gemini 3.7 Flash}
\label{app:gemini}

With Gemini 3.7 Flash (Table~\ref{tab:task_breakdown_gemini}), top-down geometry is the best addition on Viewpoint Imagination (73.0\% vs.\ 63.7\% with endpoint RGB alone), and video is the best addition on Cross-view Integration (85.3\% vs.\ 78.7\%). Adding the best representation per question type reaches 74.4\% on Viewpoint Imagination (Table~\ref{tab:per_type_selection}). Choosing the best representation for each question in hindsight reaches 82.5\% on Viewpoint Imagination and 89.0\% on Cross-view Integration (Table~\ref{tab:representation_breakdown}).

\begin{table}[!htbp]
\centering
\small
\begin{tabular}{llr}
\toprule
Task & Question Type & $n$ \\
\midrule
\multirow{7}{*}{\makecell[l]{Viewpoint\\Imagination}}
 & Open-space Direction & 91 \\
 & Nearest-obstacle Direction & 55 \\
 & Visibility & 104 \\
 & Nearest Object & 104 \\
 & Farthest Object & 98 \\
 & Object-relative Direction & 170 \\
 \cmidrule(l){2-3}
 & All & 622 \\
\midrule
\multirow{4}{*}{\makecell[l]{Cross-view\\Integration}}
 & Nearest Object & 81 \\
 & Farthest Object & 82 \\
 & Spatial Ordering & 137 \\
 \cmidrule(l){2-3}
 & All & 300 \\
\midrule
\multicolumn{2}{l}{Total} & 922 \\
\bottomrule
\end{tabular}
\caption{Number of questions per question type.}
\label{tab:question_counts}
\end{table}

\begin{table*}[!t]
\centering
\resizebox{\textwidth}{!}{%
\begin{tabular}{llrrrrrrrrr}
\toprule
& & & & \multicolumn{6}{c}{Endpoint RGB + Additional Representation} \\
\cmidrule(lr){5-10}
Task & Question Type & Observed View & + Endpoint RGB & Top-down & Text & Video & Depth & Normals & Panorama & Oracle Selection \\
\midrule
\multirow{7}{*}{\makecell[l]{Viewpoint\\Imagination}}
 & Open-space Direction & 52.7 & 58.2 & 67.0 & 63.7 & 63.7 & \underline{69.2} & 63.7 & \textbf{74.7} & 85.7 \\
 & Nearest-obstacle Direction & 36.4 & 45.5 & \textbf{61.8} & 54.5 & \underline{56.4} & 52.7 & 52.7 & \underline{56.4} & 74.5 \\
 & Visibility & 32.7 & 49.0 & \textbf{57.7} & \textbf{57.7} & 55.8 & \underline{56.7} & 51.9 & 55.8 & 69.2 \\
 & Nearest Object & 51.9 & 65.4 & \textbf{75.0} & 70.2 & 68.3 & \underline{71.2} & 69.2 & 70.2 & 80.8 \\
 & Farthest Object & 66.3 & 65.3 & \textbf{75.5} & 68.4 & 69.4 & \underline{73.5} & 69.4 & 71.4 & 80.6 \\
 & Object-relative Direction & 66.5 & 79.4 & 86.5 & 84.1 & \textbf{87.6} & 84.1 & 82.9 & \underline{87.1} & 93.5 \\
 \cmidrule(l){2-11}
 & All & 53.7 & 63.7 & \textbf{73.0} & 69.3 & 69.9 & 70.7 & 67.8 & \underline{72.0} & 82.5 \\
\midrule
\multirow{4}{*}{\makecell[l]{Cross-view\\Integration}}
 & Nearest Object & 60.5 & 75.3 & 80.2 & 80.2 & 80.2 & \textbf{81.5} & 79.0 & 80.2 & 84.0 \\
 & Farthest Object & 59.8 & 62.2 & \textbf{73.2} & 65.9 & \underline{72.0} & 68.3 & 68.3 & 69.5 & 79.3 \\
 & Spatial Ordering & 73.7 & 90.5 & 94.9 & \underline{95.6} & \textbf{96.4} & 94.9 & \underline{95.6} & 93.4 & 97.8 \\
 \cmidrule(l){2-11}
 & All & 66.3 & 78.7 & \underline{85.0} & 83.3 & \textbf{85.3} & 84.0 & 83.7 & 83.3 & 89.0 \\
\bottomrule
\end{tabular}
}
\caption{Results with Gemini 3.7 Flash using oracle representations, under the protocol of Table~\ref{tab:task_breakdown_gpt}. Oracle selection: correct if the endpoint RGB or any additional representation is correct (an upper bound). Bold and underline: best and second-best additional representation.}
\label{tab:task_breakdown_gemini}
\end{table*}

\begin{table*}[!t]
\centering
\small
\begin{tabular}{llrrrrr}
\toprule
Question Type & Added Representation & Endpoint Correct & Rescued & Total & Acc.\ (\%) & Gain (pp) \\
\midrule
Open-space Direction       & Panorama &  53 & 15 &  68/91  & 74.7 & 16.5 \\
Nearest-obstacle Direction & Top-down &  25 &  9 &  34/55  & 61.8 & 16.4 \\
Visibility                 & Text     &  51 &  9 &  60/104 & 57.7 &  8.7 \\
Nearest Object             & Top-down &  68 & 10 &  78/104 & 75.0 &  9.6 \\
Farthest Object            & Top-down &  64 & 10 &  74/98  & 75.5 & 10.2 \\
Object-relative Direction  & Video    & 135 & 14 & 149/170 & 87.6 &  8.2 \\
\midrule
All                        &          & 396 & 67 & 463/622 & 74.4 & 10.8 \\
\bottomrule
\end{tabular}
\caption{Adding the best representation per Viewpoint Imagination question type. Rescued: endpoint-RGB errors corrected by the added representation. Gain: rescued questions as a percentage of the type.}
\label{tab:per_type_selection}
\end{table*}

\begin{table}[!htbp]
\centering
\resizebox{\linewidth}{!}{%
\begin{tabular}{lrrcrrc}
\toprule
& \multicolumn{3}{c}{\makecell{Viewpoint Imagination\\($n{=}622$)}} & \multicolumn{3}{c}{\makecell{Cross-view Integration\\($n{=}300$)}} \\
\cmidrule(lr){2-4}\cmidrule(lr){5-7}
Representation & Acc. & $\Delta$ & Best For & Acc. & $\Delta$ & Best For \\
\midrule
Endpoint RGB Only     & 63.7 & --      & --  & 78.7 & --  & -- \\
\midrule
+ Top-down            & \textbf{73.0} & \imagUp{9.3} & 4/6 & 85.0 & \imagUp{6.3} & 1/3 \\
+ Text                & 69.3 & \imagUp{5.6} & 1/6 & 83.3 & \imagUp{4.7} & 0/3 \\
+ Video               & 69.9 & \imagUp{6.3} & 1/6 & \textbf{85.3} & \imagUp{6.7} & 1/3 \\
+ Depth               & 70.7 & \imagUp{7.1} & 0/6 & 84.0 & \imagUp{5.3} & 1/3 \\
+ Surface Normals     & 67.8 & \imagUp{4.2} & 0/6 & 83.7 & \imagUp{5.0} & 0/3 \\
+ Panorama            & 72.0 & \imagUp{8.4} & 1/6 & 83.3 & \imagUp{4.7} & 0/3 \\
\midrule
Oracle Selection       & 82.5 & \imagUp{18.8} & --  & 89.0 & \imagUp{10.3} & -- \\
\bottomrule
\end{tabular}
}
\caption{Representation breakdown for Gemini 3.7 Flash. $\Delta$: change from the endpoint RGB alone. Best For: number of question types on which the representation scores highest (ties counted for each). Oracle selection as in Table~\ref{tab:task_breakdown_gemini}.}
\label{tab:representation_breakdown}
\end{table}

\subsection{Implementation Details on Estimation}
\label{app:estimation}

\begin{figure*}[!t]
    \centering
    \includegraphics[width=\linewidth]{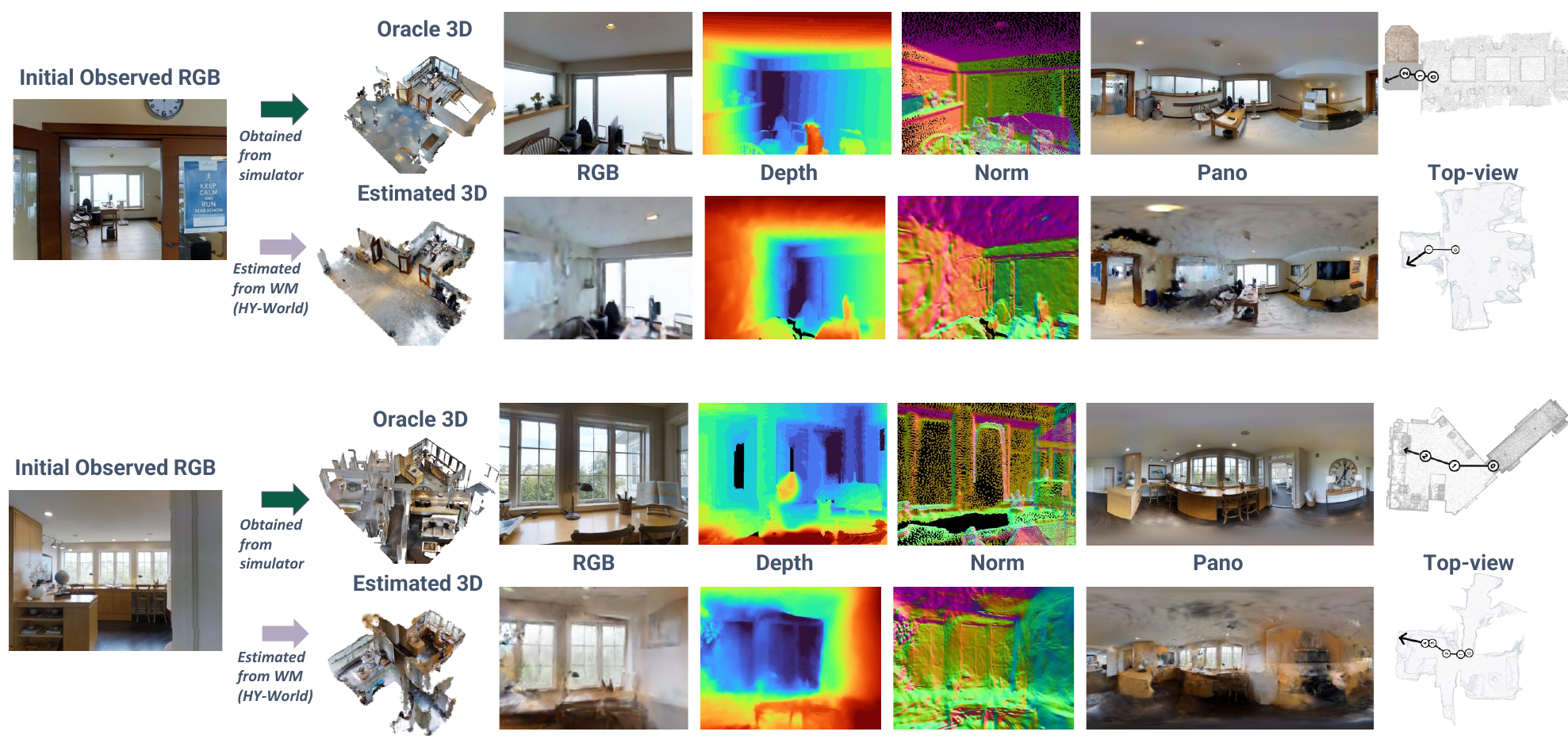}
\caption{Examples of oracle and estimated spatial representations. Oracle representations are derived from ground-truth 3D scenes; estimated representations are obtained from HY-World 2.0 for Viewpoint Imagination.}
    \label{fig:vis_example2}
\end{figure*}

For Viewpoint Imagination, we run HY-World 2.0~\citep{hyworld2026} on each initial image and export the generated world as a Gaussian-splat scene with a fused mesh. The camera trajectory is planned in this world from the initial image and the movement instruction only, without any ground-truth poses, views, or answers. Gemini 3.7 Flash decomposes the movement instruction into simple actions, describes the referenced objects, and localizes them with bounding boxes in the initial image. Ray casting from the initial camera through these image regions estimates the objects' distances and 3D positions in the generated scene. Using the scene's obstacles, we plan the shortest traversable path to each target and convert it into camera translations and rotations. The camera stops at a navigable position at a preposition-dependent distance from the target (e.g., 0.8\,m for ``walk to''). Scene scale is calibrated with metric depth estimated by MoGe-2~\citep{wang2025moge2} from the initial image. Endpoint RGB, depth, surface normals, panorama, and top-down geometry are rendered from the same generated world at the planned endpoint with the same camera settings as the oracle views. Video captures the planned camera trajectory, while spatial text describes the planned camera movements. Over all Viewpoint Imagination questions, the planned endpoints deviate from the ground truth by a median of 1.4\,m in position and $14^\circ$ in heading, which partly explains why estimated endpoint RGB falls short of oracle RGB in Sec.~\ref{sec:estimated_imagination}. Figure~\ref{fig:vis_example2} shows more examples.

For Cross-view Integration, we estimate camera poses from the observed images using VGGT-$\Omega$~\citep{wang2026vggtomega}. With two reference images, we select a target pose midway between their estimated poses. With three reference images, the model plans a camera trajectory. The reference images and estimated poses provide the context for Stable Virtual Camera (SEVA)~\citep{zhou2025stable}, which synthesizes RGB at the target viewpoint without constructing an explicit 3D scene. We use MoGe-2 (ViT-L)~\citep{wang2025moge2} to estimate depth from the synthesized endpoint RGB. Depth is the only additional estimated representation considered in this setting.

\subsection{Implementation Details of \method and Baselines}
\label{app:baseline}

\paragraph{Gating and routing.}
In the gating stage, the reasoner answers each question twice, from the observed view(s) alone and with the estimated endpoint RGB added. If the two answers agree, this answer is returned. Otherwise, the question enters the routing stage, where the reasoner sees the question, the observed view(s), and the estimated endpoint RGB, together with a short description of each option, and chooses which evidence to use without answering the question. The prompt includes brief guidance on which evidence suits each question type (prompts in Appendix~\ref{app:prompts}). The reasoner then answers using the chosen evidence. All calls use GPT-6.1-sol with high reasoning effort.

\paragraph{Ablation.}
Table~\ref{tab:gating_ablation} ablates the two stages. No Gate provides all estimated representations at once. +Gating keeps the answer when the observed-view and estimated-RGB answers agree, and otherwise falls back to No Gate. +Routing applies routing to every question. Gating+Routing is the full method, i.e., \method{} in Table~\ref{tab:wm_baselines}. On Viewpoint Imagination, each stage alone improves over No Gate, and combining them performs best.

\begin{table}[!htbp]
\centering
\small
\setlength{\tabcolsep}{4pt}
\begin{tabular}{lcccc}
\toprule
Task & No Gate & +Gating & +Routing & \makecell{Gating+\\Routing} \\
\midrule
\makecell[l]{Viewpoint\\Imagination}  & 54.7 & 58.8 & 58.5 & \textbf{61.4} \\
\makecell[l]{Cross-view\\Integration} & 77.3 & 76.3 & \textbf{77.7} & \textbf{77.7} \\
\bottomrule
\end{tabular}
\caption{Gating ablation with GPT-6.1-sol on \bench{}: overall accuracy (\%). Best per row in bold.}
\label{tab:gating_ablation}
\end{table}

\paragraph{Baselines.}
We compare MindJourney~\citep{yang2025mindjourney}, which generates imagined views for test-time spatial reasoning, and AVIC~\citep{yu2026avic}, which adaptively invokes imagination and selects trajectories. Both methods are evaluated on the same questions with GPT-6.1-sol as the final reasoner. MindJourney uses its released Search World Model and AVIC uses SEVA. 
MindJourney answers from its three most helpful imagined views and AVIC from its selected trajectory, with a note that imagined views may be inaccurate.

\subsection{Prompts of \method{}}
\label{app:prompts}

Figure~\ref{fig:prompt_answer} shows the answering prompt used for both gating predictions and the final answer. The gating stage compares the two predicted letters directly and uses no additional prompt. Figures~\ref{fig:prompt_route_vi} and~\ref{fig:prompt_route_cv} show the routing prompts. Bracketed items are placeholders for question-specific inputs.

\subsection{Use of LLMs}
We used LLMs for grammar correction, text polishing, and minor formatting suggestions.

\begin{figure*}[!tp]
\begin{tcolorbox}[colback=gray!5, colframe=gray!60, boxrule=0.5pt, arc=2pt, left=4pt, right=4pt, top=3pt, bottom=3pt,
  title=\small\textbf{Routing prompt (Viewpoint Imagination)}, fonttitle=\small]
\scriptsize
\textbf{System:} You decide what evidence to use for a spatial reasoning question. You do not answer the question.

\vspace{3pt}
\textbf{User:} Start: [START image] \; FINAL VIEW (after completing the movement): This frame shows your final position and facing direction after completing the movement in the question. Use this frame to ground your reasoning about the final state, and combine it with earlier frames to understand the movement and spatial relationships. [Estimated endpoint RGB]

Note: the FINAL VIEW above was not photographed. A world model generated a 3D scene from the START image alone, and the FINAL VIEW was rendered in that generated scene after following the movement. It may be blurry or wrong, especially for things the START image does not show.

The question the reasoner must answer (do not answer it yourself): [Question and answer options]

Answering from the START image alone and answering from the START image plus the FINAL VIEW gave DIFFERENT answers to this question, so the generated FINAL VIEW changed the conclusion and may be misleading. Choose ONE option for the final answer:
\begin{itemize}\setlength{\itemsep}{0pt}\setlength{\parsep}{0pt}\setlength{\leftmargin}{10pt}
\item start\_only: rely on the START image alone and set the generated FINAL VIEW aside (no generated evidence at all).
\item est\_rgb: use the START image together with the generated FINAL VIEW, with no additional representation.
\item topdown: an overhead map of the generated scene with the numbered positions, the path and an arrow for the final facing direction; unlabeled grey geometry, no object names or colors.
\item text: a written step-by-step description of the planned movement (distances, turns, final facing direction); no imagery.
\item video: a first-person video through the generated scene from the START view to the FINAL VIEW.
\item depth: a depth map of the generated final view (distance of each visible surface); only what the final view shows.
\item surface\_normals: a surface-normal map of the generated final view (which way each visible surface faces); only what the final view shows.
\item panorama: a 360-degree image rendered at the final position in the generated scene, showing all directions around you; directions the START image never showed are filled in by the world model.
\end{itemize}
Guidance for this kind of benchmark:
\begin{itemize}\setlength{\itemsep}{0pt}\setlength{\parsep}{0pt}\setlength{\leftmargin}{10pt}
\item Open space or the closest obstacle in some direction around you, or which object is nearest to you after moving: the panorama shows every direction at the final position, so prefer panorama.
\item Whether something can be seen after moving: the top-down map shows where things are relative to your final position and facing, so prefer topdown.
\item Which object is farthest away, or the direction of a named object relative to you: the generated view tends to mislead these answers, so prefer start\_only.
\end{itemize}
Reply with one short sentence giving your reason, then the choice.
\end{tcolorbox}
\caption{Routing prompt of \method{} for Viewpoint Imagination. The output is constrained to a JSON object with a reason and one choice.}
\label{fig:prompt_route_vi}
\end{figure*}

\begin{figure*}[!tp]
\centering
\begin{minipage}[t]{0.485\textwidth}
\vspace{0pt}
\begin{tcolorbox}[colback=gray!5, colframe=gray!60, boxrule=0.5pt, arc=2pt, left=4pt, right=4pt, top=3pt, bottom=3pt,
  title=\small\textbf{Answering prompt}, fonttitle=\small]
\scriptsize
\textbf{System (Viewpoint Imagination):} Solve the multiple-choice embodied spatial reasoning question using only the provided evidence. Images are labeled and chronological. Return only A, B, C, or D.

\vspace{3pt}
\textbf{System (Cross-view Integration):} Solve this multi-view spatial reasoning multiple-choice question using the provided evidence. Reason carefully. Return only A, B, C, or D.

\vspace{3pt}
\textbf{User:} [Observed view(s)] \; [Estimated endpoint RGB, if used] \; [Additional representation with its caption, if selected] \; [Question and answer options]
\end{tcolorbox}
\caption{Answering prompt of \method{}. The gating stage queries it without and with the estimated endpoint RGB; the final answer uses the evidence selected by routing.}
\label{fig:prompt_answer}
\end{minipage}\hfill
\begin{minipage}[t]{0.485\textwidth}
\vspace{0pt}
\begin{tcolorbox}[colback=gray!5, colframe=gray!60, boxrule=0.5pt, arc=2pt, left=4pt, right=4pt, top=3pt, bottom=3pt,
  title=\small\textbf{Routing prompt (Cross-view Integration)}, fonttitle=\small]
\scriptsize
\textbf{System:} You decide what evidence to use for a spatial reasoning question. You do not answer the question.

\vspace{3pt}
\textbf{User:} [Numbered observed images] \; [Estimated RGB]

Note: the Estimated RGB above was not photographed. It was generated as an additional view of the same scene and may be blurry or wrong in places.

The question the reasoner must answer (do not answer it yourself): [Question and answer options]

Answering from the numbered images alone and answering from the numbered images plus the Estimated RGB gave DIFFERENT answers to this question. Choose ONE option for the final answer:
\begin{itemize}\setlength{\itemsep}{0pt}\setlength{\parsep}{0pt}\setlength{\leftmargin}{10pt}
\item start\_only: rely on the numbered images alone and set the Estimated RGB aside.
\item est\_rgb: use the numbered images together with the Estimated RGB.
\item depth: use the numbered images, the Estimated RGB and an estimated depth map of the Estimated RGB (distance of each visible surface from the camera).
\end{itemize}
Guidance for this kind of benchmark:
\begin{itemize}\setlength{\itemsep}{0pt}\setlength{\parsep}{0pt}\setlength{\leftmargin}{10pt}
\item Which object is farthest from another object: distances are hard to judge from color images alone, so prefer depth.
\item Which object is nearest to another object, or the left-to-right order of objects: the Estimated RGB connects the views well, so prefer est\_rgb.
\end{itemize}
Reply with one short sentence giving your reason, then the choice.
\end{tcolorbox}
\caption{Routing prompt of \method{} for Cross-view Integration.}
\label{fig:prompt_route_cv}
\end{minipage}
\end{figure*}

\subsection{Artifact Licenses and Usage Rights}

Our work builds upon publicly available research datasets, pretrained models, and commercial language model services. We acknowledge their respective licenses and usage terms below.

\paragraph{Datasets.}
Matterport3D~\citep{chang2017matterport3d} is used to construct ImagiRep and obtain ground-truth scene geometry, camera poses, and visual observations. It is subject to the Matterport End User License Agreement for Academic Use of Model Data, which permits non-commercial academic research. The original dataset remains subject to its distribution restrictions.

\paragraph{Pretrained Models.}
HY-World 2.0~\citep{hyworld2026} is released under the Tencent HY-World 2.0 Community License Agreement, which includes geographic and usage restrictions.
Stable Virtual Camera (SEVA)~\citep{zhou2025stable} is released under the Stability AI Non-Commercial License.
MoGe-2~\citep{wang2025moge2} provides code under the MIT License, with certain third-party components under Apache 2.0.
VGGT-$\Omega$~\citep{wang2026vggtomega} is released under the FAIR Noncommercial Research License.

\paragraph{Language Models.}
GPT-6.1-sol and Gemini 3.7 Flash are accessed through their respective services and are subject to the applicable OpenAI and Google usage terms rather than open-source model licenses.
All third-party artifacts are used for academic research and evaluation. We do not redistribute third-party model weights or the original Matterport3D dataset. Their original licenses and usage restrictions remain applicable.

%% file: references.bib
@inproceedings{ray2024sat,
author = {Ray, Arijit and Duan, Jiafei and Brown, Ellis and Tan, Reuben and Bashkirova, Dina and Hendrix, Rose and Ehsani, Kiana and Kembhavi, Aniruddha and Plummer, Bryan A. and Krishna, Ranjay and Zeng, Kuo-Hao and Saenko, Kate},
booktitle = {Conference on Language Modeling (COLM)},
title = {{SAT}: Dynamic Spatial Aptitude Training for Multimodal Language Models},
year = {2025}
}

@inproceedings{wang2026anchorweave,
  title={Anchorweave: World-consistent video generation with retrieved local spatial memories},
  author={Wang, Zun and Lin, Han and Yoon, Jaehong and Cho, Jaemin and Zhang, Yue and Bansal, Mohit},
  booktitle={European Conference on Computer Vision},
  pages={203--224},
  year={2026},
  organization={Springer}
}

@inproceedings{wang2026epic,
  title={Epic: Efficient video camera control learning with precise anchor-video guidance},
  author={Wang, Zun and Cho, Jaemin and Li, Jialu and Lin, Han and Yoon, Jaehong and Zhang, Yue and Bansal, Mohit},
  booktitle={Forty-third International Conference on Machine Learning},
  year={2026}
}

@article{zhang2024spartun3d,
  title={Spartun3d: Situated spatial understanding of 3d world in large language models},
  author={Zhang, Yue and Xu, Zhiyang and Shen, Ying and Kordjamshidi, Parisa and Huang, Lifu},
  journal={arXiv preprint arXiv:2410.03878},
  year={2024}
}

@article{zhang2026seeing,
  title={Seeing Isn't Knowing: Do VLMs Know When Not to Answer Spatial Questions (and Why)?},
  author={Zhang, Yue and Wang, Zun and Lin, Han and Bitton, Yonatan and Szpektor, Idan and Bansal, Mohit},
  journal={arXiv preprint arXiv:2605.30557},
  year={2026}
}

@article{agarwal2025cosmos,
  title={Cosmos world foundation model platform for physical {AI}},
  author={Agarwal, Niket and Ali, Arslan and Bala, Maciej and Balaji, Yogesh and Barker, Erik and Cai, Tiffany and Chattopadhyay, Prithvijit and Chen, Yongxin and Cui, Yin and Ding, Yifan and others},
  journal={arXiv preprint arXiv:2501.03575},
  year={2025}
}

@article{tversky2003structures,
  title={Structures of mental spaces: How people think about space},
  author={Tversky, Barbara},
  journal={Environment and behavior},
  volume={35},
  number={1},
  pages={66--80},
  year={2003},
  publisher={Sage Publications}
}

@article{burgess2006spatial,
  title={Spatial memory: how egocentric and allocentric combine},
  author={Burgess, Neil},
  journal={Trends in cognitive sciences},
  volume={10},
  number={12},
  pages={551--557},
  year={2006},
  publisher={Elsevier}
}

@article{rieser1989access,
  title={Access to knowledge of spatial structure at novel points of observation.},
  author={Rieser, John J},
  journal={Journal of Experimental Psychology: Learning, Memory, and Cognition},
  volume={15},
  number={6},
  pages={1157},
  year={1989},
  publisher={American Psychological Association}
}

@inproceedings{zhou2025stable,
  title={{Stable Virtual Camera}: Generative view synthesis with diffusion models},
  author={Zhou, Jensen and Gao, Hang and Voleti, Vikram and Vasishta, Aaryaman and Yao, Chun-Han and Boss, Mark and Torr, Philip and Rupprecht, Christian and Jampani, Varun},
  booktitle={2025 IEEE/CVF International Conference on Computer Vision (ICCV)},
  pages={12405--12414},
  year={2025},
  organization={IEEE}
}

@inproceedings{bar2025navigation,
  title={Navigation world models},
  author={Bar, Amir and Zhou, Gaoyue and Tran, Danny and Darrell, Trevor and LeCun, Yann},
  booktitle={2025 IEEE/CVF Conference on Computer Vision and Pattern Recognition (CVPR)},
  pages={15791--15801},
  year={2025},
  organization={IEEE}
}

@article{du2023learning,
  title={Learning universal policies via text-guided video generation},
  author={Du, Yilun and Yang, Sherry and Dai, Bo and Dai, Hanjun and Nachum, Ofir and Tenenbaum, Josh and Schuurmans, Dale and Abbeel, Pieter},
  journal={Advances in neural information processing systems},
  volume={36},
  pages={9156--9172},
  year={2023}
}

@inproceedings{xu2025mmsibench,
author = {Yang, Sihan and Xu, Runsen and Xie, Yiman and Yang, Sizhe and Li, Mo and Lin, Jingli and Zhu, Chenming and Chen, Xiaochen and Duan, Haodong and Yue, Xiangyu and Lin, Dahua and Wang, Tai and Pang, Jiangmiao},
booktitle = {International Conference on Learning Representations (ICLR)},
title = {{MMSI-Bench}: A Benchmark for Multi-Image Spatial Intelligence},
year = {2026}
}

@article{yuan2026fast,
  title={{Fast-WAM}: Do world action models need test-time future imagination?},
  author={Yuan, Tianyuan and Dong, Zibin and Liu, Yicheng and Zhao, Hang},
  journal={arXiv preprint arXiv:2603.16666},
  year={2026}
}

@article{zhu2026thinking,
  title={Thinking with Imagination: Agentic Visual Spatial Reasoning with World Simulators},
  author={Zhu, Chenming and Lin, Jingli and Long, Yilin and Cao, Peizhou and Wang, Tai and Pang, Jiangmiao and Liu, Xihui},
  journal={arXiv preprint arXiv:2606.06476},
  year={2026}
}

@inproceedings{mindcube,
author = {Wang, Qineng and Yin, Baiqiao and Zhang, Pingyue and Zhang, Jianshu and Wang, Kangrui and Wang, Zihan and Zhang, Jieyu and Chandrasegaran, Keshigeyan and Liu, Han and Krishna, Ranjay and Xie, Saining and Wu, Jiajun and Fei-Fei, Li and Li, Manling},
booktitle = {International Conference on Learning Representations},
title = {{MindCube}: Spatial Mental Modeling from Limited Views},
year = {2026}
}

@inproceedings{allangles,
author = {Yeh, Chun-Hsiao and Wang, Chenyu and Tong, Shengbang and Cheng, Ta-Ying and Wang, Ruoyu and Chu, Tianzhe and Zhai, Yuexiang and Chen, Yubei and Gao, Shenghua and Ma, Yi},
booktitle = {Proceedings of the AAAI Conference on Artificial Intelligence},
doi = {10.1609/aaai.v40i14.38188},
pages = {12000--12008},
title = {Seeing from Another Perspective: Evaluating Multi-View Understanding in {MLLMs}},
volume = {40},
year = {2026}
}

@inproceedings{li2026sawbench,
author = {Li, Chuhan and Han, Rilyn and Hsu, Joy and Liang, Yongyuan and Dhawan, Rajiv and Wu, Jiajun and Yang, Ming-Hsuan and Wang, Xin Eric},
booktitle = {Forty-third International Conference on Machine Learning},
title = {{SAW-Bench}: Learning Situated Awareness in the Real World},
year = {2026}
}

@inproceedings{huang2025vista,
author = {Huang, Yanjia and Wu, Mingyang and Li, Renjie and Tu, Zhengzhong},
booktitle = {2026 IEEE International Conference on Robotics and Automation},
title = {{VISTA}: Generative Visual Imagination for Vision-and-Language Navigation},
year = {2026}
}

@inproceedings{yang2025mindjourney,
author = {Yang, Yuncong and Liu, Jiageng and Zhang, Zheyuan and Zhou, Siyuan and Tan, Reuben and Yang, Jianwei and Du, Yilun and Gan, Chuang},
booktitle = {Advances in Neural Information Processing Systems},
doi = {10.52202/085713-3669},
pages = {121747--121777},
publisher = {Curran Associates, Inc.},
title = {{MindJourney}: Test-Time Scaling with World Models for Spatial Reasoning},
volume = {38},
year = {2025}
}

@article{lu2024generative,
  title={{Generative World Explorer}},
  author={Lu, Taiming and Shu, Tianmin and Yuille, Alan and Khashabi, Daniel and Chen, Jieneng},
  journal={arXiv preprint arXiv:2411.11844},
  year={2024}
}

@inproceedings{zhu2025aether,
  title     = {Aether: Geometric-Aware Unified World Modeling},
  author    = {{Aether Team} and Zhu, Haoyi and Wang, Yifan and Zhou, Jianjun and Chang, Wenzheng and Zhou, Yang and Li, Zizun and Chen, Junyi and Shen, Chunhua and Pang, Jiangmiao and He, Tong},
  booktitle = {ICCV},
  year      = {2025}
}

@article{yu2026avic,
author = {Yu, Shoubin and Zhang, Yue and Wang, Zun and Yoon, Jaehong and Yao, Huaxiu and Ding, Mingyu and Bansal, Mohit},
journal = {arXiv preprint arXiv:2602.08236},
title = {When and How Much to Imagine: Adaptive Test-Time Scaling with World Models for Visual Spatial Reasoning},
year = {2026}
}

@inproceedings{hwang2026spatio,
author = {Hwang, Chan Yeong and Choi, Miso and On, Sunghyun and Kim, Jinkyu and Lee, Jungbeom},
booktitle = {European Conference on Computer Vision},
title = {{SpatiO}: Adaptive Test-Time Orchestration of Vision-Language Agents for Spatial Reasoning},
year = {2026}
}

@inproceedings{chang2017matterport3d,
author = {Chang, Angel and Dai, Angela and Funkhouser, Thomas and Halber, Maciej and Nie{\ss}ner, Matthias and Savva, Manolis and Song, Shuran and Zeng, Andy and Zhang, Yinda},
booktitle = {International Conference on 3D Vision (3DV)},
title = {{Matterport3D}: Learning from {RGB-D} Data in Indoor Environments},
year = {2017}
}

@inproceedings{yang2026cambrian,
  title={{Cambrian-S}: Towards spatial supersensing in video},
  author={Yang, Shusheng and Yang, Jihan and Huang, Pinzhi and Brown, Ellis and Yang, Zihao and Yu, Yue and Tong, Shengbang and Zheng, Zihan and Xu, Yifan and Wang, Muhan and others},
  booktitle={International Conference on Learning Representations},
  year={2026}
}

@article{parker2025genie,
  title={Genie 3: A new frontier for world models},
  author={Parker-Holder, Jack and Fruchter, Shlomi},
  journal={Google DeepMind Blog},
  year={2025}
}

@inproceedings{li2026viewspatial,
  title={{ViewSpatial-Bench}: Evaluating multi-perspective spatial localization in vision-language models},
  author={Li, Dingming and Li, Hongxing and Wang, Zixuan and Yan, Yuchen and Zhang, Hang and Chen, Siqi and Hou, Guiyang and Jiang, Shengpei and Zhang, Wenqi and Shen, Yongliang and others},
  booktitle={European Conference on Computer Vision},
  year={2026},
  organization={Springer}
}

@inproceedings{ma20253dsrbench,
  title={{3DSRBench}: A comprehensive 3d spatial reasoning benchmark},
  author={Ma, Wufei and Chen, Haoyu and Zhang, Guofeng and Chou, Yu-Cheng and Chen, Jieneng and De Melo, Celso and Yuille, Alan},
  booktitle={2025 IEEE/CVF International Conference on Computer Vision (ICCV)},
  pages={6924--6934},
  year={2025},
  organization={IEEE}
}

@inproceedings{chen2024spatialvlm,
  title={{SpatialVLM}: Endowing vision-language models with spatial reasoning capabilities},
  author={Chen, Boyuan and Xu, Zhuo and Kirmani, Sean and Ichter, Brian and Sadigh, Dorsa and Guibas, Leonidas and Xia, Fei},
  booktitle={Proceedings of the IEEE/CVF conference on computer vision and pattern recognition},
  pages={14455--14465},
  year={2024}
}

@inproceedings{fu2024blink,
  title={{BLINK}: Multimodal large language models can see but not perceive},
  author={Fu, Xingyu and Hu, Yushi and Li, Bangzheng and Feng, Yu and Wang, Haoyu and Lin, Xudong and Roth, Dan and Smith, Noah A and Ma, Wei-Chiu and Krishna, Ranjay},
  booktitle={European Conference on Computer Vision},
  pages={148--166},
  year={2024},
  organization={Springer}
}

@inproceedings{yang2025thinking,
  title={Thinking in space: How multimodal large language models see, remember, and recall spaces},
  author={Yang, Jihan and Yang, Shusheng and Gupta, Anjali W and Han, Rilyn and Fei-Fei, Li and Xie, Saining},
  booktitle={2025 IEEE/CVF Conference on Computer Vision and Pattern Recognition (CVPR)},
  pages={10632--10643},
  year={2025},
  organization={IEEE}
}

@inproceedings{wang2025moge2,
  title     = {{MoGe-2}: Accurate Monocular Geometry with Metric Scale and Sharp Details},
  author    = {Wang, Ruicheng and Xu, Sicheng and Dong, Yue and Deng, Yu and Xiang, Jianfeng and Lv, Zelong and Sun, Guangzhong and Tong, Xin and Yang, Jiaolong},
  booktitle = {Advances in Neural Information Processing Systems},
  volume    = {38},
  year      = {2025},
}

@article{hyworld2026,
  title={{HY-World 2.0}: A Multi-Modal World Model for Reconstructing, Generating, and Simulating {3D} Worlds},
  author={{Team HY-World}},
  journal={arXiv preprint arXiv:2604.14268},
  year={2026}
}

@inproceedings{wang2026vggtomega,
  title={{VGGT-$\Omega$}},
  author={Wang, Jianyuan and Chen, Minghao and Zhang, Shangzhan and Karaev, Nikita and Sch{\"o}nberger, Johannes and Labatut, Patrick and Bojanowski, Piotr and Novotny, David and Vedaldi, Andrea and Rupprecht, Christian},
  booktitle={Proceedings of the IEEE/CVF Conference on Computer Vision and Pattern Recognition (CVPR)},
  year={2026}
}

@misc{openai2026gpt61sol,
  author       = {{OpenAI}},
  title        = {Addendum to {GPT-6} {Astra} System Card: {GPT-6.1} {Sol}},
  howpublished = {OpenAI Deployment Safety Hub},
  year         = {2026},
  month        = sep,
  url          = {https://deploymentsafety.openai.com/gpt-6-1-sol}
}

@misc{googledeepmind2026gemini37flash,
  author       = {{Google DeepMind}},
  title        = {{Gemini 3.7 Flash} Model Card},
  howpublished = {Google DeepMind},
  year         = {2026},
  month        = aug,
  url          = {https://deepmind.google/models/model-cards/gemini-3-7-flash/}
}
